# Asynchronous Bidirectional Decoding for Neural Machine Translation

Xiangwen Zhang[1], Jinsong Su[1]*, Yue Qin[1], Yang Liu[2], Rongrong Ji[1], Hongji Wang[1]

Xiamen University, Xiamen, China[1]
Tsinghua University, Beijing, China[2]
xwzhang@stu.xmu.edu.cn, jssu@xmu.edu.cn, qinyue@stu.xmu.edu.cn
liuyang2011@tsinghua.edu.cn, rrji@xmu.edu.cn, hjw@xmu.edu.cn

## Abstract

The dominant neural machine translation (NMT) models apply unified attentional encoder-decoder neural networks for translation. Traditionally, the NMT decoders adopt recurrent neural networks (RNNs) to perform translation in a left-to-right manner, leaving the target-side contexts generated from right to left unexploited during translation. In this paper, we equip the conventional attentional encoder-decoder NMT framework with a backward decoder, in order to explore bidirectional decoding for NMT. Attending to the hidden state sequence produced by the encoder, our backward decoder first learns to generate the target-side hidden state sequence from right to left. Then, the forward decoder performs translation in the forward direction, while in each translation prediction timestep, it simultaneously applies two attention models to consider the source-side and reverse target-side hidden states, respectively. With this new architecture, our model is able to fully exploit source- and target-side contexts to improve translation quality altogether. Experimental results on NIST Chinese-English and WMT English-German translation tasks demonstrate that our model achieves substantial improvements over the conventional NMT by 3.14 and 1.38 BLEU points, respectively. The source code of this work can be obtained from https://github.com/DeepLearnXMU/ABD-NMT.

## Introduction

Recently, end-to-end neural machine translation (NMT) (Kalchbrenner and Blunsom 2013; Sutskever, Vinyals, and Le 2014; Cho et al. 2014) has achieved promising results and gained increasing attention. Compared with conventional statistical machine translation (SMT) (Koehn, Och, and Marcu 2003; Chiang 2007) which needs to explicitly design features to capture translation regularities, NMT aims to construct a unified encoder-decoder framework based on neural networks to model the entire translation process. Further, the introduction of the attention mechanism (Bahdanau, Cho, and Bengio 2015) enhances the capability of NMT in capturing long-distance dependencies. Despite being a relatively new framework, the attentional encoder-decoder NMT quickly become the de facto method.

*Corresponding author.

| Source | rì fángwèitīng zhǎngguān : bú wàng jūnguó lìshǐ zūnzhòng línguó zūnyán |
|---|---|
| Reference | *japan defense chief : never forget militaristic history , respect neighboring nations ' dignity* |
| L2R | *japan 's defense agency chief : death of militarism respects its neighbors ' dignity* |
| R2L | *japanese defense agency has never forgotten militarism 's history to respect the dignity of neighboring countries* |

Table 1: Translation examples of NMT systems with different decoding manners. **L2R/R2L** denotes the translation produced by the NMT system with left-to-right/right-to-left decoding. Texts highlighted in wavy/dashed lines are incorrect/correct translations, respectively.

Generally, most NMT decoders are based on recurrent neural networks (RNNs) and generate translations in a left-to-right manner. Thus, despite the advantage of encoding unbounded target words predicted previously for the prediction at each time step, these decoders are incapable of capturing the reverse target-side context for translation. Once errors occur in previous predictions, the quality of subsequent predictions would be undermined due to the negative impact of the noisy forward encoded target-side contexts. Intuitively, the reverse target-side contexts are also crucial for translation predictions, since they not only provide complementary signals but also bring different biases to NMT model (Hoang, Haffari, and Cohn 2017). Take the example in Table 1 into consideration. The latter half of the Chinese sentence, misinterpreted by the conventional NMT system, is accurately translated by the NMT system with right-to-left decoding. Therefore, it is important to investigate how to integrate reverse target-side contexts into the decoder to improve translation performance of NMT.

To this end, many researchers resorted to introducing bidirectional decoding into NMT (Liu et al. 2016; Sennrich, Haddow, and Birch 2016a; Hoang, Haffari, and Cohn 2017). Most of them re-ranked candidate translations using bidirectional decoding scores together, in order to select a translation with both proper prefixes and suffixes. However, such methods also come with some drawbacks limiting the potential of bidirectional decoding in NMT. On the one hand, due to the limited search space and search errors of beam search, the generated 1-best translation is often far from satisfactory and thus it fails to provide sufficient information as a complement for the other decoder. On the other hand, because the bidirectional decoders are often independent from each other during the translation, the unidirectional decoder is unable to fully exploit target-side contexts produced by the other decoder, and consequently the generated candidate translations are still undesirable. Therefore, how to effectively exert the influence of bidirectional decoding on NMT is still worthy of further study.

In this paper, we significantly extend the conventional attentional encoder-decoder NMT framework by introducing a backward decoder, for the purpose of fully exploiting reverse target-side contexts to improve NMT. As shown in Fig. 1, along with our novel asynchronous bidirectional decoders, the proposed model remains an end-to-end attentional NMT framework, which mainly consists of three components: 1) an encoder embedding the input source sentence into bidirectional hidden states; 2) a backward decoder that is similar to the conventional NMT decoder but performs translation in the right-to-left manner, where the generated hidden states encode the reverse target-side contexts; 3) a forward decoder that generates the final translation from left to right and introduces two attention models simultaneously considering the source-side bidirectional and target-side reverse hidden state vectors for translation prediction. Compared with the previous related NMT models, our model has the following advantages: 1) The backward decoder learns to produce hidden state vectors that essentially encode semantics of potential hypotheses, allowing the following forward decoder to utilize richer target-side contexts for translation. 2) By integrating right-to-left target-side context modeling and left-to-right translation generation into an end-to-end joint framework, our model alleviates the error propagation of reverse target-side context modeling to some extent.

The major contributions of this paper are concluded as follows:

- We thoroughly analyze and point out the existing drawbacks of researches on NMT with bidirectional decoding.
- We introduce a backward decoder to encode the left-to-right target-side contexts, as a supplement to the conventional context modeling mechanism of NMT. To the best of our knowledge, this is the first attempt to investigate the effectiveness of the end-to-end attentional NMT model with asynchronous bidirectional decoders.
- Experiments on Chinese-English and English-German translation show that our model achieves significant improvements over the conventional NMT model.

## Our Model

As described above, our model mainly includes three components: 1) a neural encoder with parameter set $\theta_e$; 2) a neural backward decoder with parameter set $\theta_b$; and 3) a neural forward decoder with parameter set $\theta_f$, which will be elaborated in the following subsections.

Particularly, we choose Gated Recurrent Unit (GRU) (Cho et al. 2014) to build the encoder and decoders, as it is widely used in the NMT literature with relatively few parameters required. However, it should be noted that our model is also applicable to other RNNs, such as Long Short-Term Memory (LSTM) (Hochreiter and Schmidhuber 1997).

### The Neural Encoder

The neural encoder of our model is identical to that of the dominant NMT model, which is modeled using a bidirectional RNN.

The forward RNN reads a source sentence $\mathbf{x}=x_1, x_2...x_N$ in a left-to-right order. At each timestep, we apply a recurrent activation function $\phi(\cdot)$ to learn the semantic representation of the word sequence $\mathbf{x}_{1:i}$ as $\overrightarrow{h}_i = \phi(\overrightarrow{h}_{i-1}, x_i)$. Likewise, the backward RNN scans the source sentence in the reverse order and generates the semantic representation $\overleftarrow{h}_i$ of the word sequence $\mathbf{x}_{i:N}$. Finally, we concatenate the hidden states of these two RNNs to form an annotation sequence $\mathbf{h} = \{h_1, h_2, ...h_i..., h_N\}$, where $h_i = [\overrightarrow{h}_i^T, \overleftarrow{h}_i^T]^T$ encodes information about the $i$-th word with respect to all the other surrounding words in the source sentence.

In our model, these annotations will provide source-side contexts for not only the forward decoder but also the backward one via different attention models.

### The Neural Backward Decoder

The neural backward decoder of our model is also similar to the decoder of the dominant NMT model, while the only difference is that it performs decoding in a right-to-left way.

Given the source-side hidden state vectors of the encoder and all target words generated previously, the backward decoder models how to reversely produce the next target word. Using this decoder, we calculate the conditional probability of the reverse translation $\overleftarrow{\mathbf{y}} = (y_0, y_1, y_2, ..., y_{M'})$ as follows

$$P(\overleftarrow{\mathbf{y}}|\mathbf{x}; \theta_e, \theta_b) = \prod_{j'=0}^{M'} P(y_{j'}|\mathbf{y}_{>j'}, \mathbf{x}; \theta_e, \theta_b)$$

$$= \prod_{j'=0}^{M'} \overleftarrow{g}(y_{j'+1}, \overleftarrow{s}_{j'}, m_{j'}^{eb}), \quad (1)$$

where $\overleftarrow{g}(\cdot)$ is a non-linear function, $\overleftarrow{s}_{j'}$ and $m_{j'}^{eb}$ denote the decoding state and the source-side context vector at the $j'$-th time step, respectively, and $M'$ indicates the length of the reverse translation.

Among $\overleftarrow{s}_{j'}$ and $m_{j'}^{eb}$, $\overleftarrow{s}_{j'}$ is computed by the GRU activation function $f(\cdot)$: $\overleftarrow{s}_{j'} = f(\overleftarrow{s}_{j'+1}, y_{j'+1}, m_{j'}^{eb})$, and $m_{j'}^{eb}$ is defined by a **encoder-backward decoder attention model** as the weighted sum of the source annotations $\{h_i\}$:

$$m_{j'}^{eb} = \sum_{i=1}^{N} \alpha_{j',i}^{eb} \cdot h_i, \quad (2)$$

$$\alpha_{j',i}^{eb} = \frac{exp(e_{j',i}^{eb})}{\sum_{i'=1}^{N} exp(e_{j',i'}^{eb})}, \quad (3)$$

$$e_{j',i}^{eb} = (v_a^{eb})^T tanh(W_a^{eb} \overleftarrow{s}_{j'+1} + U_a^{eb} h_i), \quad (4)$$

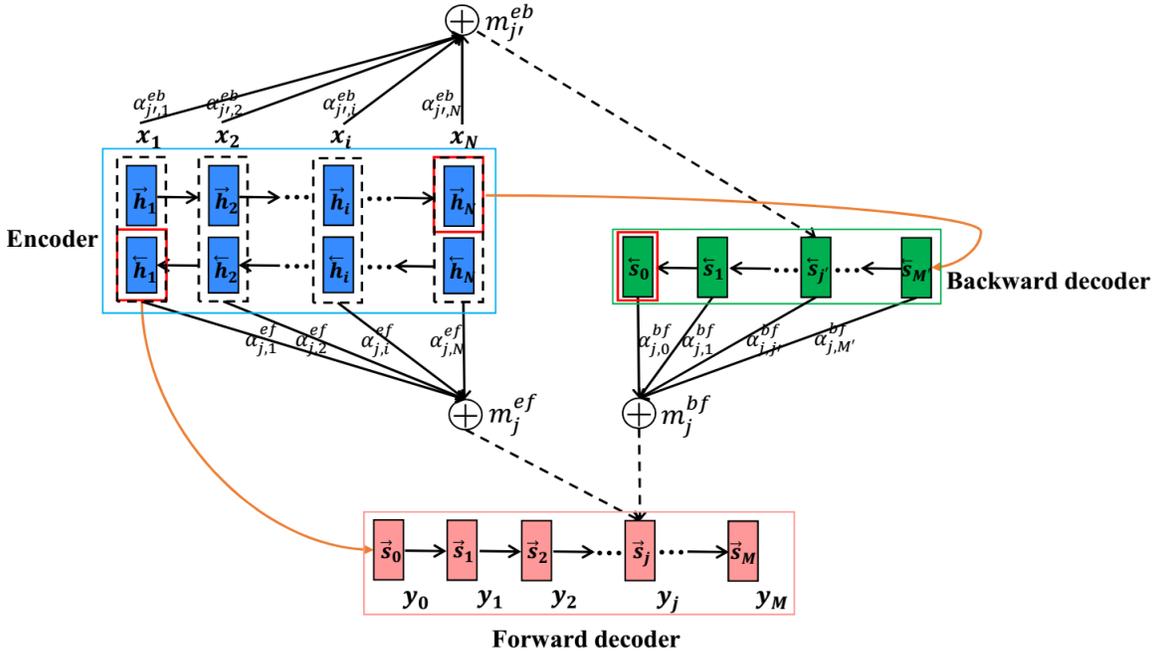

Figure 1: The architecture of the proposed NMT model. Note that the forward decoder directly attends to the reverse hidden state sequence $\overleftarrow{s}=\{\overleftarrow{s}_0,\overleftarrow{s}_1,...\overleftarrow{s}_{M'}\}$ rather than the word sequence produced by the backward decoder.

where $v_a^{eb}$, $W_a^{eb}$ and $U_a^{eb}$ are the parameters of the encoder-backward decoder attention model. In doing so, the decoder is also able to automatically select the effective source words to reversely predict target words.

By introducing this backward decoder, our NMT model is able to better exploit target-side contexts for translation prediction. In addition to the generation of target word sequence, more importantly, our backward decoder will produce target-side hidden states $\overleftarrow{s}$, which essentially captures richer reverse target-side contexts for the further use of the forward decoder.

**The Neural Forward Decoder**

The neural forward decoder of our model is extended from the decoder of the dominant NMT model. It performs decoding in a left-to-right manner under the semantic guides of source-side and reverse target-side contexts, which are separately captured by the encoder and the backward decoder.

The forward decoder is trained to sequentially predict the next target word given the source-side hidden state vectors of the encoder, the reverse target-side hidden state sequence generated by the backward encoder, and all target words generated previously. Formally, the conditional probability of the translation $\mathbf{y}=(y_0, y_1, ..., y_M)$ is defined as follows:

$$P(\mathbf{y}|\mathbf{x};\boldsymbol{\theta_e},\boldsymbol{\theta_b},\boldsymbol{\theta_f}) = \prod_{j=0}^{M} P(y_j|\mathbf{y}_{<j},\mathbf{x};\boldsymbol{\theta_e},\boldsymbol{\theta_b},\boldsymbol{\theta_f})$$

$$= \prod_{j=0}^{M} g(y_{j-1}, s_j, m_j^{ef}, m_j^{bf}), \quad (5)$$

where $g(\cdot)$ is a non-linear function, $s_j$ is the decoding state, $m_j^{ef}$ and $m_j^{bf}$ denote the source-side and reverse target-side context vectors at the $j$-th timestep, respectively.

As illustrated in Fig. 1, we use the first hidden state of the reverse encoder, denoted as $\overleftarrow{h}_1$, to initialize the first hidden state $s_0$ of the forward decoder. More importantly, we introduce two attention models to respectively capture the source-side and reverse target-side contexts: one is the **encoder-forward decoder attention model** that focuses on the source annotations and the other is the **backward decoder-forward decoder attention model** considering all reverse target-side hidden states. Specifically, we produce $m_j^{ef}$ from the hidden states $\{h_i\}$ of the encoder as follows:

$$m_j^{ef} = \sum_{i=1}^{N} \alpha_{j,i}^{ef} \cdot h_i, \quad (6)$$

$$\alpha_{j,i}^{ef} = \frac{exp(e_{j,i}^{ef})}{\sum_{i'=1}^{N} exp(e_{j,i'}^{ef})}, \quad (7)$$

$$e_{j,i}^{ef} = (v_a^{ef})^T tanh(W_a^{ef} s_{j-1} + U_a^{ef} h_i), \quad (8)$$

where $v_a^{ef}$, $W_a^{ef}$, and $U_a^{ef}$ are the parameters of the encoder-forward decoder attention model. Note that we directly choose hidden state sequence rather than word sequence to model the target-side contexts, for the reason that the former enables our model to better avoid negative effect of translation prediction errors to some extent. Likewise, we define $m_j^{bf}$ as a weighted sum of the hidden states $\{\overleftarrow{s}_{j'}\}$ of the

backward decoder:

$$m_j^{bf} = \sum_{j'=0}^{M'} \alpha_{j,j'}^{bf} \cdot \overleftarrow{s}_{j'}, \quad (9)$$

$$\alpha_{j,j'}^{bf} = \frac{exp(e_{j,j'}^{bf})}{\sum_{j''=1}^{M'} exp(e_{j,j''}^{bf})}, \quad (10)$$

$$e_{j,j'}^{bf} = (v_a^{bf})^T tanh(W_a^{bf} s_{j-1} + U_a^{bf} \overleftarrow{s}_{j'}), \quad (11)$$

where $v_a^{bf}$, $W_a^{bf}$, and $U_a^{bf}$ are the parameters of the backward decoder-forward decoder attention model.

Then, we incorporate $m_j^{ef}$ and $m_j^{bf}$ into the GRU hidden unit of the forward decoder. Formally, the hidden state $s_j$ of the forward decoder is computed by

$$\begin{aligned} s_j &= (1-z_j^d) \circ s_{j-1} + z_j^d \circ \tilde{s}_j, \\ \tilde{s}_j &= tanh(W^d v(y_{j-1}) + U^d [r_j^d \circ s_{j-1}] \\ &\quad + C^{ef} m_j^{ef} + C^{bf} m_j^{bf}), \end{aligned} \quad (12)$$

where $W^d$, $U^d$, $C^{ef}$, and $C^{bf}$ are the weight matrices, $z_j^d$ and $r_j^d$ are update and reset gates of GRU, respectively, depending on $y_{j-1}$, $s_{j-1}$, $m_j^{ef}$ and $m_j^{bf}$.

Finally, we further define the probability of $y_j$ as

$$p(y_j|\mathbf{y}_{<j},\mathbf{x};\boldsymbol{\theta_e},\boldsymbol{\theta_b},\boldsymbol{\theta_f}) \propto exp(g(y_{j-1}, s_j, m_j^{ef}, m_j^{bf})), \quad (13)$$

where $y_{j-1}$, $s_j$, $m_j^{ef}$ and $m_j^{bf}$ are concatenated and fed through a single feed-forward layer.

### Training and Testing

Given a training corpus $D=\{(\mathbf{x},\mathbf{y})\}$, we train the proposed model according to the following objective:

$$J(D;\boldsymbol{\theta_e},\boldsymbol{\theta_b},\boldsymbol{\theta_f}) = \frac{1}{|D|} \underset{\boldsymbol{\theta_e},\boldsymbol{\theta_b},\boldsymbol{\theta_f}}{\arg\max} \sum_{(\mathbf{x},\mathbf{y})\in D} \quad (14)$$

$$\{\lambda \cdot logP(\mathbf{y}|\mathbf{x};\boldsymbol{\theta_e},\boldsymbol{\theta_b},\boldsymbol{\theta_f}) + (1-\lambda) \cdot logP(\overleftarrow{\mathbf{y}}|\mathbf{x};\boldsymbol{\theta_e},\boldsymbol{\theta_b})\}$$

where $\overleftarrow{\mathbf{y}}$ is obtained by inverting $\mathbf{y}$, and $\lambda$ is a hyper-parameter used to balance the preference between the two terms.

The first term $logP(\mathbf{y}|\mathbf{x};\boldsymbol{\theta_e},\boldsymbol{\theta_b},\boldsymbol{\theta_f})$ models the translation procedure illustrated in Figure 1. To ensure the consistency between model training and testing, we perform beam search to generate reverse hidden states $\overleftarrow{s}$ when optimizing $logP(\mathbf{y}|\mathbf{x};\boldsymbol{\theta_e},\boldsymbol{\theta_b},\boldsymbol{\theta_f})$. In addition, to guarantee the $\overleftarrow{s}$ produced by beam search is of high quality, we further introduce the second term $logP(\overleftarrow{\mathbf{y}}|\mathbf{x};\boldsymbol{\theta_e},\boldsymbol{\theta_b})\}$ to maximize the conditional likelihood of $\overleftarrow{\mathbf{y}}$. Note that the beam search requires high time complexity, and therefore, we directly adopt greedy search to implement right-to-left decoding, while proves to be sufficiently effective in our experiments.

Once the proposed model is trained, we adopt a two-phase scheme to translate the unseen input sentence $\mathbf{x}$: First, we use the backward decoder with greedy search to sequentially generate $\overleftarrow{s}$ until the target-side start symbol $\langle s \rangle$ occurs with the highest probability. Then, we perform beam search on the forward decoder to find the best translation that approximately maximizes $logP(\mathbf{y}|\mathbf{x};\boldsymbol{\theta_e},\boldsymbol{\theta_b},\boldsymbol{\theta_f})$.

## Experiments

We evaluated the proposed model on NIST Chinese-English and WMT English-German translation tasks.

### Setup

For Chinese-English translation, the training data consists of 1.25M bilingual sentences with 27.9M Chinese words and 34.5M English words. These sentence pairs are mainly extracted from LDC2002E18, LDC2003E07, LDC2003E14, Hansards portion of LDC2004T07, LDC2004T08 and LDC2005T06. We chose NIST 2002 (MT02) dataset as our development set, and the NIST 2003 (MT03), 2004 (MT04), 2005 (MT05), and 2006 (MT06) datasets as our test sets. Finally, we evaluated the translations using BLEU (Papineni et al. 2002).

For English-German translation, we used WMT 2015 training data that contains 4.46M sentence pairs with 116.1M English words and 108.9M German words. Particularly, we segmented words via byte pair encoding (BPE) (Sennrich, Haddow, and Birch 2016b). The news-test 2013 was used as development set and the news-test 2015 as test set.

To efficiently train NMT models, we trained each model with sentences of length up to 50 words. In doing so, 90.12% and 89.03% of the Chinese-English and English-German parallel sentences were covered in the experiments. Besides, we set the vocabulary size to 30K for Chinese-English translation, and 50K for English-German translation, and mapped all the out-of-vocabulary words in the Chinese-English corpus to a special token UNK. Finally, such vocabularies contained 97.4% Chinese words and 99.3% English words of the Chinese-English corpus, and almost 100.0% English words and 98.2% German words of the English-German corpus, respectively. We applied *Rmsprop* (Graves 2013) (momentum = 0, $\rho$ = 0.95, and $\epsilon$ = 1× $10^{-4}$) to train models for 5 epochs and selected the best model parameters according to the model performance on the development set. During this procedure, we set the following hyper-parameters: word embedding dimension as 620, hidden layer size as 1000, learning rate as $5 \times 10^{-4}$, batch size as 80, gradient norm as 1.0, and dropout rate as 0.3. All the other settings are the same as in (Bahdanau, Cho, and Bengio 2015).

### Baselines

We compared the proposed model against the following state-of-the-art SMT and NMT systems:

- *Moses*[1]: an open source phrase-based translation system with default configuration and a 4-gram language model trained on the target portion of training data. Note that we used all data to train MOSES.

- *RNNSearch*: a re-implementation of the attention-based NMT system (Bahdanau, Cho, and Bengio 2015) with slight changes from dl4mt tutorial[2].

- *RNNSearch(R2L)*: a variant of RNNSearch that produces translation in a right-to-left direction.

---
[1]http://www.statmt.org/moses/
[2]https://github.com/nyu-dl/dl4mt-tutorial

| SYSTEM | MT03 | MT04 | MT05 | MT06 | Average |
|---|---|---|---|---|---|
| COVERAGE | 34.49 | 38.34 | 34.91 | 34.25 | 35.50 |
| MemDec | 36.16 | 39.81 | 35.91 | 35.98 | 36.97 |
| DeepLAU | 39.35 | 41.15 | 38.07 | 37.29 | 38.97 |
| DMAtten | 38.33 | 40.11 | 36.71 | 35.29 | 37.61 |
| Moses | 32.93 | 34.76 | 31.31 | 31.05 | 32.51 |
| RNNSearch | 36.59 | 39.57 | 35.56 | 35.29 | 36.75 |
| RNNSearch(R2L) | 36.54 | 39.70 | 35.61 | 34.67 | 36.63 |
| ATNMT | 38.09 | 40.99 | 36.87 | 36.17 | 38.03 |
| NSC(RT) | 37.68 | 40.82 | 36.21 | 35.50 | 37.55 |
| NSC(HS) | 37.99 | 40.74 | 36.82 | 36.32 | 37.97 |
| Our Model | **40.02** | **42.32** | **38.84** | **38.38** | **39.89** |

Table 2: Evaluation of the NIST Chinese-English translation task using case-insensitive BLEU scores ($\lambda$=0.7). Here we displayed the experimental results of the first four models reported in (Wang et al. 2017; Zhang et al. 2017). COVERAGE (Tu et al. 2016) is a basic NMT model with a coverage model. MemDec (Wang et al. 2016) improves translation quality with external memory. DeepLAU (Wang et al. 2017) reduces the gradient propagation length inside the recurrent unit of RNN-based NMT. DMAtten (Zhang et al. 2017) incorporates word reordering knowledge into attentional NMT.

- *ATNMT*: an attention-based NMT system with two directional decoders (Liu et al. 2016) which explores the **agreement on target-bidirectional NMT**. Using this model, we first run beam search for forward and backward models independently to obtain two $k$-best lists, and then re-score the combination of these two lists using the joint model to find the best candidate. Following (Liu et al. 2016), we set both beam sizes of two decoders as 10. Note that we replaced LSTM adopted in (Liu et al. 2016) with GRU to ensure fair comparison.

- *NSC(RT)*: it is a variant of **neural system combination** framework proposed by Zhou et al. (2017). It first uses an attentional NMT model consisting of one standard encoder and one backward decoder to produce the best **reverse translation**. Finally, another attentional NMT model generates the final output from its standard encoder and a reverse translation encoder which embeds the best reverse translation, in a way similar to the multi-source NMT model (Zoph and Knight 2016). This model differs from ours in two aspects: (1) it is not an end-to-end model, and (2) it considers the embedded hidden states of the reverse translation, while our model considers the hidden states produced by the backward decoder.

- *NSC(HS)*: it is similar to NSC(RT), with the only difference that it directly considers the reverse **hidden states** produced by the backward decoder.

We set beam sizes of all above-mentioned models as 10, and the beam sizes of the backward and forward decoders of our model as 1 and 10, respectively.

### Results on Chinese-English Translation

**Parameters.** RNNSearch, RNNSearch(R2L), ATNMT, NSC(RT), NSC(HS) models have 85.6M, 85.6M, 171.2M, 120.0M and 130.0M parameters, respectively. By contrast, the parameter size of our model is about 130.0M.

**Speed.** We used a single GPU device 1080Ti to train models. It takes one hour to train 6,500, 6,500, 6,500 and 4,700 and 3,708 minibatches for RNNSearch, RNNSearch(R2L), ATNMT, NSC(RT), NSC(HS) models, respectively. The training speed of the proposed model is relatively slow: about 1,758 mini-batches are processed in one hour.

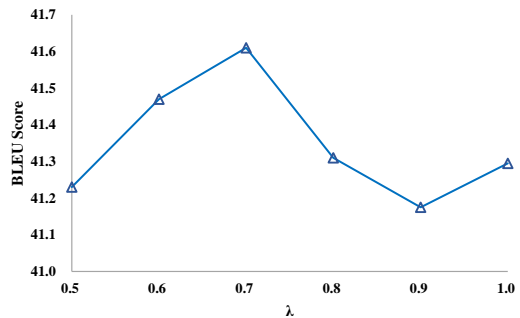

Figure 2: Experiment results on the development set using different $\lambda$s.

We first investigated the impact of the hyper-parameter $\lambda$ (see Eq. (14)) on the development set. To this end, we gradually varied $\lambda$ from 0.5 to 1.0 with an increment of 0.1 in each step. As shown in Fig. 2, we find that our model achieved the best performance when $\lambda$=0.7. Therefore, we set $\lambda$=0.7 for all experiments thereafter.

The experimental results on Chinese-English translation are depicted in Table 2. We also displayed the performances of some dominant individual models such as ***COVERAGE*** (Tu et al. 2016), ***MemDec*** (Wang et al. 2016), ***DeepLAU*** (Wang et al. 2017) and ***DMAtten*** (Zhang et al. 2017) on the same data set. Specifically, the proposed model significantly outperforms Moses, RNNSearch, RNNSearch(R2L), ATNMT, NSC(RT) and NSC(HS) by 7.38, 3.14, 3.26, 1.86, 2.34, and 1.92 BLEU points, respectively. Even when compared with (Tu et al. 2016; Wang et al. 2016; 2017; Zhang et al. 2017), our model still has better performance in the same setting. Moreover, we draw the following conclusions:

(1) In contrast to RNNSearch and RNNSearch(R2L), our model exhibits much better performance. These results testify our hypothesis that the forward and backward decoders are complementary to each other in target-side context modeling, and therefore, the simultaneous exploration of bidirectional decoders will lead to better translations.

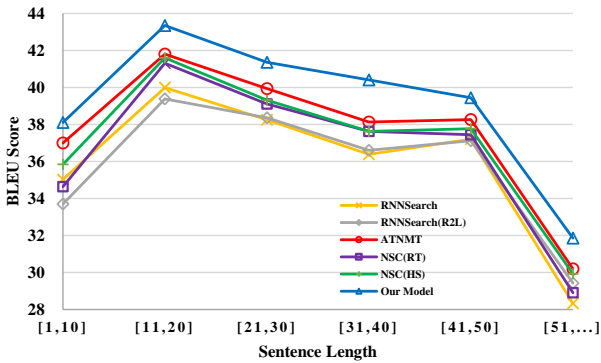

Figure 3: BLEU scores on different translation groups divided according to source sentence length.

| SYSTEM | TEST |
|---|---|
| BPEChar | 23.90 |
| RecAtten | 25.00 |
| ConvEncoder | 24.20 |
| Moses | 20.54 |
| RNNSearch | 24.88 |
| RNNSearch(R2L) | 23.83 |
| ATNMT | 25.08 |
| NSC(RT) | 25.15 |
| NSC(HS) | 25.36 |
| Our Model | **26.26** |

Table 4: Evaluation of the WMT English-German translation task using case-sensitive BLEU scores ($\lambda$=0.8). We directly cited the experimental results of the first three models provided by (Gehring et al. 2017). BPEChar (Chung, Cho, and Bengio 2016) is an attentional NMT model with a character-level decoder. RecAtten (Yang et al. 2017) uses a recurrent attention model to explicitly model the dependence between attentions among target words. ConvEncoder (Gehring et al. 2017) introduces a convolutional encoder into NMT.

(2) On all test sets, our model outperforms ATNMT, which indicates that compared with $k$-best hypotheses rescoring (Liu et al. 2016), joint modeling with attending to reverse hidden states behaves better in exploiting reverse target-side contexts. The underlying reason is that the reverse hidden states encode richer target-side contexts than single translation. In addition, compared with the $k$-best hypotheses rescoring, our model could refine translation at a more fine-grained level via the attention mechanism.

(3) Particularly, the fact that NSC(HS) outperforms NSC(RT) reveals the advantage of reverse hidden state representations of the backward decoder in overcoming data sparsity. Besides, our model behaves better than NSC(HS), which accords with our intuition that to some extent, joint model is able to alleviate the error propagation when encoding target-side contexts.

(4) Note that the performance of our model is better than that of our model (RR). This result verifies our speculation that model training with the translations obtained by greedy search is superior due to the consistency during the training and testing procedure.

Finally, based on the length of source sentences, we divided our test sets into different groups and then compared the system performances in each group. Fig. 3 illustrates the BLEU scores on these groups of test sets. We observe that our model achieves the best performance in all groups, although the performances of all systems drop with the increase of the length of source sentences. These results clearly demonstrate once again the effectiveness of our model.

### Case Study

To better understand how our model outperforms others, we studied the 1-best translations using different models.

Table 3 provides a Chinese-English translation example. We find that RNNSearch produces the translation with good prefix, while RNNSearch(R2L) generates the translation with desirable suffix. Although there are various models with bidirectional decoding that could exploit bidirectional contexts, most of them are unable to translate the whole sentence precisely and our model is currently the only one capable to produce a high quality translation in this circumstance.

### Results on English-German Translation

To enhance the persuasion of our experiments, we also provided some experiments results on the same data set, including ***BPEChar*** (Chung, Cho, and Bengio 2016), ***RecAtten*** (Yang et al. 2017), and ***ConvEncoder*** (Gehring et al. 2017). We determined the optimal $\lambda$ as 0.8 according to the performance of our model on the development set.

Table 4 presents the results on English-German translation. Our model still significantly outperforms others including some dominant NMT systems with other improved techniques. We believe that our work can be applied to other architectures easily. It should be noted that the BLEU score gaps between our model and the others on English-German translation are much smaller than those on Chinese-English translation. The underlying reasons lie in the following two aspects, which have also been mentioned in (Shen et al. 2016). First, the Chinese-English datasets contain four reference translations for each sentence while the English-German dataset only have single reference. Second, compared with German, Chinese is more distantly related to English, leading to the predominant advantage of utilizing target-side contexts in Chinese-English translation.

## Related Work

In this work, we mainly focus on how to exploit bidirectional decoding to refine translation, which has always been a research focus in machine translation.

In SMT, many approaches through backward language model (BLM) or target-bidirectional decoding have been explored to capture right-to-left target-side contexts for translation. For example, Watanabe and Sumita (2002) explored two decoding methods: one is the right-to-left decoding based on the left-to-right beam search algorithm; the other decodes in both directions and merges the two hypothesized partial sentences into one. Finch and Sumita (2009) integrated both mono-directional approaches to reduce the effects caused by language specificity. Particularly, they integrated the BLM to their reverse translation decoder. Beyond left-to-right decoding, Zhang et al. (2013) studied the effects of multiple decomposition structures as well as dynamic bidirectional decomposition on SMT.

| | |
|---|---|
| **Source** | yīyuè kāishǐ , zǒngwùshěng jiāng yǒu liù míng zhíyuán yī zhōu zhìshǎo yī tiān bù xūyào jìn bàngōngshì , kěyǐ zài jiā lǐ , dàxué huò túshūguǎn tòuguò gāosù wǎngluò fúwù gōngzuò . |
| **Reference** | starting from january , the ministry of internal affairs and communications will have six employees who do n't need to go to their offices at least one day a week ; instead they may work from home , universities or libraries through high - speed internet services . |
| **Moses** | since january , there will be six staff members – a week for at least one day in office , they can at home , university or through high - speed internet library services . |
| **RNNSearch** | as early as january , six staff members will not be required to enter office at least one day in one week , which can be done through high - speed internet services through high - speed internet services . |
| **RNNSearch(R2L)** | beginning in january , at least six staff members have to go to the office for at least one week and can work at home , and university or library through high - speed internet services . |
| **ATNMT** | at the beginning of january , there will be six staff members to go to office at least one week , which can be done through high - speed internet services at home and university or libraries . |
| **NSC(RT)** | at least six staff members will leave office for at least one week at least one week , and can work at home and university or library through high - speed internet services . |
| **NSC(HS)** | in january , there will be six staff members who are required to enter offices for at least one day at least one day , and we can work at home , university or library through high - speed internet services . |
| **Our Model** | starting in january , six staff members will not need to enter the office at least one day in one week, and they can work at home , universities or libraries through high - speed internet services . |

Table 3: Translation examples of different systems. Texts highlighted in wavy lines are incorrectly translated. Please note that the translations produced by RNNSearch and RNNSearch(R2L) are complementary to each other, and the translation generated by our model is the most accurate and complete.

When it comes to NMT, the dominant RNN-based NMT models also perform translation in a left-to-right manner, leading to the same drawback of underutilization of target-side contexts. To address this issue, Liu et al. (2016) first jointly train both directional LSTM models, and then in testing they try to search for target-side translations which are supported by both models. Similarly, Sennrich et al. (2016a) attempted to re-rank the left-to-right decoding results by right-to-left decoding, leading to diversified translation results. Recently, Hoang et al. (2017) proposed an approximate inference framework based on continuous optimization that enables decoding bidirectional translation models. Finally, it is noteworthy that our work is also related to pre-translation (Niehues et al. 2016; Zhou et al. 2017) and neural automatic post-editing (Pal et al. 2017; Dowmunt and Grundkiewicz 2017) for NMT, because our model involves two stages of translation.

Overall, the most relevant models include (Liu et al. 2016; Sennrich, Haddow, and Birch 2016a; Hoang, Haffari, and Cohn 2017; Zhou et al. 2017; Pal et al. 2017; Dowmunt and Grundkiewicz 2017). Our model significantly differs from these works in the following aspects: 1) The motivation of our work varies from theirs. Specifically, in this work, we aim to fully exploit the reverse target-side contexts encoded by right-to-left hidden state vectors to improve NMT with left-to-right decoding. In contrast, Liu et al. (2016), Sennrich et al. (2016a), Hoang et al. (2017) investigated how to exploit bidirectional decoding scores to produce better translations, both Niehues et al. (2016) and Zhou et al. (2017) intended to combine the advantages of both NMT and SMT, and in the work of (Pal et al. 2017; Dowmunt and Grundkiewicz 2017), they explored multiple neural architectures for the task of automatic post-editing of machine translation output. 2) Our model attends to right-to-left hidden state vectors, while (Niehues et al. 2016; Zhou et al. 2017; Pal et al. 2017; Dowmunt and Grundkiewicz 2017) considered the raw best output of machine translation system instead. 3) Our model is an end-to-end NMT model, while the bidirectional decoders adopted in (Liu et al. 2016; Sennrich, Haddow, and Birch 2016a; Hoang, Haffari, and Cohn 2017) were independent from each other, and the component used to produce the raw translation was independent from the NMT model in (Niehues et al. 2016; Zhou et al. 2017; Pal et al. 2017; Dowmunt and Grundkiewicz 2017). In addition, Serban et al. (2018) introduced a backward RNN to refine the forward RNN. But their work was only applied in the training procedure, which is different from ours where both the forward and backward RNNs were simultaneously used for sequence generation.

## Conclusions and Future Work

In this paper, we have equipped the conventional attentional encoder-decoder NMT model with a backward decoder. In our model, the backward decoder first produces hidden state vectors encoding reverse target-side contexts. Then, two individual hidden state sequences generated by the encoder and the backward decoder are simultaneously exploited via attention mechanism by the forward decoder for translation.

Compared with the previous models, ours is an end-to-end NMT model that fully utilizes reverse target-side contexts for translation. Experimental results on Chinese-English and English-German translation tasks demonstrate the effectiveness of our model.

Our model is generally applicable to other models with RNN-based decoder. Therefore, the effectiveness of our approach on other tasks related to RNN-based decoder modeling, such as image captioning, will be investigated in future research. Moreover, in our work, the attention mechanisms acting on the encoder and the backward decoder are independent from each other. However, intuitively, these two mechanisms should be closely associated with each other. Therefore, we are interested in exploring better attention mechanism combination to further refine our model.

## Acknowledgments

The authors were supported by National Natural Science Foundation of China (Nos. 61672440, 61573294 and 61432013), Scientific Research Project of National Language Committee of China (Grant No. YB135-49), Natural Science Foundation of Fujian Province of China (No. 2016J05161), and National Key R&D Program of China (Nos. 2017YFC011300 and 2016YFB1001503). We also thank the reviewers for their insightful comments.

## References

Bahdanau, D.; Cho, K.; and Bengio, Y. 2015. Neural machine translation by jointly learning to align and translate. In *Proc. of ICLR2015*.

Chiang, D. 2007. Hierarchical phrase-based translation. *Computational Linguistics* 33:201–228.

Cho, K.; van Merrienboer, B.; Gulcehre, C.; Bahdanau, D.; Bougares, F.; Schwenk, H.; and Bengio, Y. 2014. Learning phrase representations using rnn encoder–decoder for statistical machine translation. In *Proc. of EMNLP2014*, 1724–1734.

Chung, J.; Cho, K.; and Bengio, Y. 2016. A character-level decoder without explicit segmentation for neural machine translation. In *Proc. of ACL2016*, 1693–1703.

Dowmunt, M. J., and Grundkiewicz, R. 2017. An exploration of neural sequence-to-sequence architectures for automatic post-editing. In *arXiv:1706.04138v1*.

Finch, A., and Sumita, E. 2009. Bidirectional phrase-based statistical machine translation. In *Proc. of EMNLP2009*, 1124–1132.

Gehring, J.; Auli, M.; Grangier, D.; and Dauphin, Y. 2017. A convolutional encoder model for neural machine translation. In *Proc. of ACL2017*, 123–135.

Graves, A. 2013. Generating sequences with recurrent neural networks. In *arXiv:1308.0850v5*.

Hoang, C. D. V.; Haffari, G.; and Cohn, T. 2017. Decoding as continuous optimization in neural machine translation. In *arXiv1701.02854*.

Hochreiter, S., and Schmidhuber, J. 1997. Long short-term memory. *Neural Computation* 1735–1780.

Kalchbrenner, N., and Blunsom, P. 2013. Recurrent continuous translation models. In *Proc. of EMNLP2013*, 1700–1709.

Koehn, P.; Och, F. J.; and Marcu, D. 2003. Statistical phrase-based translation. In *Proc. of NAACL2003*, 48–54.

Liu, L.; Utiyama, M.; Finch, A.; and Sumita, E. 2016. Agreement on target-bidirectional neural machine translation. In *Proc. of NAACL2016*, 411–416.

Niehues, J.; Cho, E.; Ha, T.-L.; and Waibel, A. 2016. Pre-translation for neural machine translation. In *Proc. of COLING2016*, 1828–1836.

Pal, S.; Naskar, S. K.; Vela, M.; Liu, Q.; and van Genabith, J. 2017. Neural automatic post-editing using prior alignment and reranking. In *Proc. of EACL2017*, 349–355.

Papineni, K.; Roukos, S.; Ward, T.; and Zhu, W. 2002. Bleu: A method for automatic evaluation of machine translation. In *Proc. of ACL2002*, 311–318.

Sennrich, R.; Haddow, B.; and Birch, A. 2016a. Edinburgh neural machine translation systems for wmt 16. In *arXiv:1606.02891v2*.

Sennrich, R.; Haddow, B.; and Birch, A. 2016b. Neural machine translation of rare words with subword units. In *Proc. of ACL2016*, 1715–1725.

Serban, I. V.; Sordoni, A.; Bengio, Y.; Courville, A.; and Pineau, J. 2018. Twin networks: Matching the future for sequence generation. In *Proc. of ICLR2018*.

Shen, S.; Cheng, Y.; He, Z.; He, W.; Wu, H.; Sun, M.; and Liu, Y. 2016. Minimum risk training for neural machine translation. In *Proc. of ACL2016*, 1683–1692.

Sutskever, I.; Vinyals, O.; and Le, Q. V. 2014. Sequence to sequence learning with neural networks. In *Proc. of NIPS2014*, 3104–3112.

Tu, Z.; Lu, Z.; Liu, Y.; Liu, X.; and Li, H. 2016. Modeling coverage for neural machine translation. In *Proc. of ACL2016*, 76–85.

Wang, M.; Lu, Z.; Li, H.; and Liu, Q. 2016. Memory-enhanced decoder for neural machine translation. In *Proc. of EMNLP2016*, 278–286.

Wang, M.; Lu, Z.; Zhou, J.; and Liu, Q. 2017. Deep neural machine translation with linear associative unit. In *Proc. of ACL2017*, 136–145.

Watanabe, T., and Sumita, E. 2002. Bidirectional decoding for statistical machine translation. In *Proc. of COLING 2002*, 1200–1208.

Yang, Z.; Hu, Z.; Deng, Y.; Dyer, C.; and Smola, A. 2017. Neural machine translation with recurrent attention modeling. In *Proc. of EACL2017*, 383–387.

Zhang, H.; Toutanova, K.; Quirk, C.; and Gao, J. 2013. Beyond left-to-right: Multiple decomposition structures for smt. In *Proc. of NAACL2013*, 12–21.

Zhang, J.; Wang, M.; Liu, Q.; and Zhou, J. 2017. Incorporating word reordering knowledge into attention-based neural machine translation. In *Proc. of ACL 2017*, 1524–1534.

Zhou, L.; Hu, W.; Zhang, J.; and Zong, C. 2017. Neural system combination for machine translation. In *Proc. of ACL2017*, 378–384.

Zoph, B., and Knight, K. 2016. Multi-source neural translation. In *Proc. of NAACL2016*, 30–34.